\documentclass[conference]{IEEEtran}
\ifCLASSINFOpdf
\else
\fi

\usepackage{times}
\usepackage{epsfig}
\usepackage{graphicx}
\usepackage{amsmath}
\usepackage{amssymb}
\usepackage{booktabs}
\usepackage{epstopdf}
\DeclareGraphicsExtensions{.eps}
\usepackage{algorithm}%
\usepackage{algpseudocode}%

\begin{document}
%
\title{Mining Mid-level Features for Action Recognition Based on Effective Skeleton Representation}

\author{\IEEEauthorblockN{Pichao Wang, 
Wanqing Li,
Philip Ogunbona and
Zhimin Gao}
\IEEEauthorblockA{University of Wollongong, Wollongong, NSW, Australia, 2522\\
pw212@uowmail.edu.au, \{wanqing, philipo\}@uow.edu.au, \{zg126\}@uowmail.edu.au }
\IEEEauthorblockN{Hanling Zhang}
\IEEEauthorblockA{Hunan University, P. R. China, jt\_hlzhang@hnu.edu.cn}
}

\maketitle

\begin{abstract}
Recently, mid-level features have shown promising performance in computer vision. Mid-level features learned by incorporating class-level information are potentially more discriminative than traditional low-level local features. In this paper, an effective method is proposed to extract mid-level features from Kinect skeletons for 3D human action recognition. Firstly, the orientations of limbs connected by two skeleton joints are computed and each orientation is encoded into one of the 27 states indicating the spatial relationship of the joints. Secondly, limbs are combined into parts and the limb's states are mapped into part states. Finally, frequent pattern mining is employed to mine the most frequent and relevant (discriminative, representative and non-redundant) states of parts in continuous several frames. These parts are referred to as \textit{Frequent Local Parts} or \textit{FLPs}. The \textit{FLPs} allow us to build powerful \textit{bag-of-FLP}-based action representation. This new representation yields state-of-the-art results on MSR DailyActivity3D and MSR ActionPairs3D.     
\end{abstract}

\section{Introduction}
Human action recognition has been an active research topic in computer vision due to its wide range of applications, such as smart surveillance and human-computer interactions. Despite remarkable research efforts and encouraging advances in the past decade, accurate recognition of human actions is still an open problem.

A common and intuitive method to represent human motion is to use a sequence of skeletons. With the development of the cost-effective depth cameras and algorithms for real-time pose estimation \cite{Shotton2011}, skeleton extraction has become more and more robust and skeleton-based action representation is becoming one of the most practical and promising approaches. Up to date, the skeleton-based approach primarily focuses on low-level features and models the dynamics of the skeletons holistically, such as moving pose \cite{Zanfir_movpose} and trajectories of human joints \cite{Gowayyed2013_HOD}. The full skeletal description is highly subject to the noise introduced during the extraction of the skeleton and less effective in the cases where some actions involve motion of the whole body and others are preformed using only small number of body parts. A key fact we observed is that during the temporal axis of actions, only a few body parts in several continuous  frames are activated during the performance of the actions. These parts are more robust and discriminative to represent an action. In our method we take advantage of this observation to capture mid-level features for action recognition.

Inspired by the mid-level features mining techniques \cite{fernando2014mining} for image classification, we propose a new scheme applying pattern mining to obtain the most relevant combinations of parts in several continuous  frames for action recognition rather than to utilize all the joints as most previous works did. In particular,  a new descriptor called \textit{bag-of-FLPs} is proposed to describe an action as illustrated in Fig. 1.   
\begin{figure}[!ht]
\begin{center}
{\includegraphics[height = 130mm, width = 90mm]{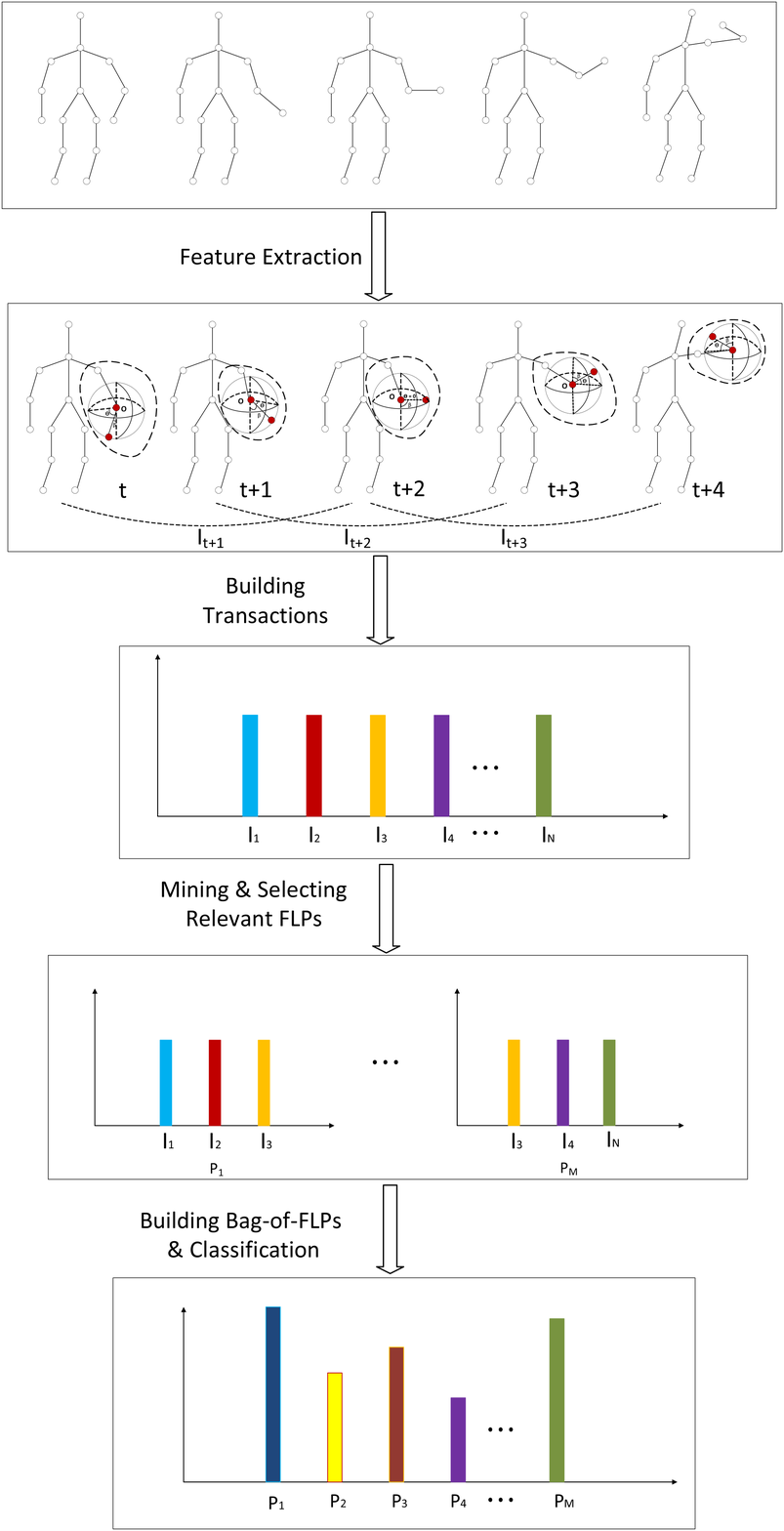}}
\end{center}
\caption{The general framework of the proposed method.}
\label{fig:framework}
\end{figure}
The overall process of our method can be divided into four steps: feature extraction, building transactions, mining \& selecting relevant patterns and building \textit{Bag-of-FLPs} \& classification. We first compute the orientations of limbs, i.e. connected joints, and then encode each orientation into one of the 27 states indicating the spatial relationship of the joints. Limbs are combined into parts and limb's states are mapped to part states. Local temporal information is included by combining part states of several, say, 5, continuous frames into one transaction for mining, with each state as one item. In order to keep motion information after frequent pattern mining, the unique states of parts of the continuous frames are reserved, removing the repeated ones, ensuring the pose information and motion information be included in each transaction. The most relevant patterns, which we referred to \textit{FLPs}, are mined and selected o represent frames and build \textit{bag-of-FLPs} as new representation for a whole action. The new representation is much robust to the errors in the features, because the errors are usually not frequent patterns. 

Our main contributions include the following four aspects. First, an effective and efficient method is proposed to extract skeleton features. Second, a novel method is developed to explore spatial and temporal information in skeleton data, simultaneously. Third, an effective scheme is proposed for applying pattern mining to action recognition by adapting the generic pattern mining tools to the features of skeleton. Our scheme is much robust to noise as most noisy data does not form frequent patterns. In addition, our scheme has achieved the state-of-the-art results on several benchmark datasets. 
 
The rest of the paper is organized as follows. Section II reviews the related work. Section III presents our scheme in detail. Section IV shows experimental results. Conclusion is made in section V.

\section{Related work}
The process of action recognition can be generally divided into two main steps, action representation and action classification. Action representation consists of feature extraction and feature selection. Features can be extracted from input sources such as depth maps, skeleton and/or RGB images. Regardless of the input source, there are two main approaches, space-time approach and sequential approach \cite{Li2008,Li2010,ye2013survey}, to the representation of actions. The space-time approach usually extracts local or holistic features from space-time volume, without explicit modelling of temporal dynamics. By contrast, the sequential approach normally extracts local features from each frame of the input source and models the dynamics explicitly. Action classification is the step of learning a classifier based on action representation and classifying any new observations using the classifier. For space-time approaches, discriminative classifier, such as Support Vector Machine (SVM), is often used for classification. For the sequential approach, generative statistical models, such as Hidden Markov Model (HMM), are commonly used. Our method belongs to the skeleton-based space-time volume approach. In this section, we mainly review the existing work of skeleton-based action representation for action recognition. 

For the skeleton-based sequential approach, Xia et al. \cite{xia2012view} proposed a feature called Histograms of 3D Joint Locations (HOJ3D) as a representation of postures. The HOJ3D essentially encodes spatial occupancy information relative to the root joint, e.g. hip centre. A modified spherical coordinate system is defined on the root joint and the 3D space is divided into $N$ bins. The HOJ3D is reprojected using Linear Discriminant Analysis (LDA) to reduce dimensionality and then clustered into $K$ posture visual words which represent the prototypical poses of actions. HMMs are adopted to model the visual words and recognize actions. Radial distance is adopted in this spherical coordinate system which makes the method to some extend view-invariant. 

Koppula et al. \cite{Koppula2013} explicitly modelled the motion hierarchy to enable their method to handle simple human-object interactions. The human activities and object affordances are jointly modelled as a Markov Random Field (MRF) where the nodes represent objects and sub-activities, and the edges represent the relationships between object affordances, their relations with sub-activities, and their evolution over time. Feature vectors that represent the object's location and the changing information in the scene are defined by training a Structural Support Vector Machine (SSVM). Similar to this approach, Sung et al. \cite{sung2012unstructured} proposed a hierarchical two-layer Maximum Entropy Markov Model (MEMM) to represent an activity. The lower layer nodes represent sub-activities while higher level nodes describe more complex activities, for example, ``lifting left hand" and ``pouring water" can be described as a sub-activity and a complex activity, respectively. Wang et al. \cite{wang2014learning} proposed an Local Occupancy Patterns (LOP) feature calculated from the 3D point cloud around a particular joint to discriminate different types of interactions and Fourier Temporal Pyramid (FTP) to represent the temporal structure. Based on above two types of features, a model called Actionlet Ensemble Model (AEM) is proposed which is a combination of the features for a subset of the joints. Due to the numerous actionlets, data mining technique is used to discover discriminative actionlets. Both skeleton and point cloud information are utilized to recognize human-objects interactions. 

For the skeleton-based space-time volume approach, Yang et al. \cite{Yang2012} proposed a new feature descriptor called EigenJoints features which contain posture features, motion features and offset features. The pair-wise joint differences in current frames and their consecutive frames are used to encode the spatial and temporal information, which are called posture features and motion features, respectively. The difference of a pose with respect to the initial pose is called offset features. The initial pose is generally assumed as a neutral pose. The three channels are normalized and  Principal Component Analysis (PCA) is applied to reduce redundancy and noise to obtain the EigenJoints descriptor. A Naive-Bayes-Nearest-Neighbor (NBNN) classifier is adopted to recognize actions. Gowayyed et al. \cite{Gowayyed2013_HOD} proposed a new descriptor called Histograms of Oriented Displacements (HOD) to recognize actions. The displacement of each joint votes with its length in a histogram of oriented angles. Each 3D trajectory is represented by the HOD of its three 2D projection. In order to reserve temporal information, a temporal pyramid is proposed, where trajectories are considered as a whole, halves and quarters and then all the descriptors in these three levels are concatenated to form the final descriptor. A linear SVM is used to classify actions based on the histograms. Similar to this work, Hussein et al. \cite{Hussein2013} proposed a descriptor called Covariance of 3D Joints (Cov3DJ) for human action recognition. This descriptor uses covariance matrix to capture the dependence of locations of different joints on one another during an action. In order to capture the order of motion in time, a hierarchy of Cov3DJs is used, similarly to the work in \cite{Gowayyed2013_HOD}. 

Zanfir et al. \cite{Zanfir_movpose} proposed a descriptor called moving pose which is formed by the position, velocity and acceleration of skeleton joints within a short time window around the current fame. To learn discriminative pose, a modified $k$-Nearest Neighbours ($k$NN) classifier is used that considers both the temporal location of a particular frame within the action sequence as well as the discrimination power of its moving pose descriptor compared to other frames in the training set. Thanh et al. \cite{Thanh2014}
extracted key frames which are the central frames in the short temporal segments of videos and labelled each key frame as a pattern for a unit action. An improved Term Frequency-Inverse Document Frequency (TF-IDF) method is used to learn the discriminative patterns and learned patterns is defined as local features for action recognition. Wang et al. \cite{wang2013approach} first estimated human joints positions from videos and then grouped the estimated joints into five parts. Each action is represented by computing sets of co-occurring spatial and temporal configurations of body parts. The authors use a bag of words method with the extracted features for classification. Ohn-Bar and Trivedi \cite{Bar2013} tracked the joint angles and built a descriptor based on similarities between angle trajectories. This feature is further combined with a double-HOG descriptor that accounts for the spatio-temporal distribution of depth values around the joints. Theodorakopoulos et al. \cite{Theodorakopoulos201412} initially processed the skeleton data from sensor coordinate to torso PCA frame in order to gain robust and invariant pose representation. Sparse coding in dissimilarity space is utilized to sparsely represent the actions. Chaaraoui et al. \cite{Chaaraoui2014786} proposed to use an evolutionary algorithm to determine the optimal subset of skeleton joints, taking into account the topological structure of the skeleton. 

To fuse depth-based features with skeleton-based features, Althloothi et al. \cite{Althloothi2014} presented two sets of features, features for shape representation extracted from depth data by using a spherical harmonics representation and features for kinematic structure extracted from skeleton data by estimating 3D joint positions. The shape features are used to describe the 3D silhouette structure while the kinematic features are used to describe the movement of the human body. Both sets of features are fused at the kernel level for action recognition by using Multiple Kernel Learning (MKL) technique. Similar to this direction, Chaaraoui et al. \cite{Chaaraoui2013} proposed a fusion method to combine skeleton and silhouette-based features. The skeletal features are obtained by normalising the 3D position of original skeleton data while the silhouette-based features are generated by extracting contour points of the silhouette. After feature fusion, a model called bag of key poses is employed for action recognition. The key poses are obtained by $K$-means clustering algorithm and the words are made up of key poses. In recognition stage, unknown video sequences are classified based on sequence matching. Rahmani et al. \cite{rahmanireal} proposed an algorithm combining the discriminative information from depth maps as well as from 3D joints positions for action recognition. To avoid the suppression of subtle discriminative information, local information integration and normalization are performed. Joint importance is encoded by using joint motion volume. Random Decision Forest (RDF) is trained to select the discriminant features. Because of the low dimensionality of their features, their method turns to be efficient. 

In above methods, most of them are based low-level features and need the whole skeletal description which leads to their weak adaptation to noise. In addition, most of them need to explore the spatial and temporal information, separately, and then combine them together. Besides, most of the methods used to explore temporal information are subject to the neural poses, which are shared by all actions. However, in our method, we use a parts-based mid-level feature to represent actions and explore the spatial and temporal information simultaneously. This makes our method more robust.  

\section{Proposed Method}
The overall process of the proposed method is illustrated in Fig. 1. It can be divided into four steps: feature extraction, building transactions, mining \& selecting relevant patterns and building \textit{Bag-of-FLPs} \& classification.
\subsection{Feature Extraction}
In our method, the orientations of human body limbs are considered as low-level features and they can be calculated from the two joints of the limbs. For Kinect skeleton data, $20$ joint positions, as shown in Fig. 2, are tracked \cite{Shotton2011}. The skeleton data is first normalized using Algorithm 1 in \cite{Zanfir_movpose} to suppress noise in the original skeleton data and to compensate for length variations across different subjects and different body parts. Each joint $i$ has 3 coordinates, denoted as $(x_{i}, y_{i}, z_{i})$ after normalization.
\begin{figure}[!ht]
\begin{center}
{\includegraphics[height = 50mm, width = 30mm]{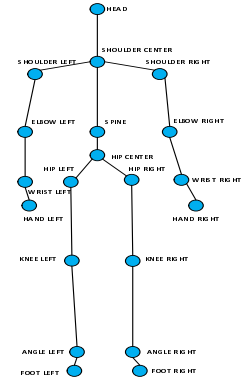}}
\end{center}
\caption{The human joints tracked with the skeleton tracker \cite{Shotton2011}.}
\label{fig:framework}
\end{figure}

For Kinect skeleton, it is found that the \textit{Hand Left}, \textit{Hand Right}, \textit{Foot Left}, \textit{Foot Right}, and \textit{Spine} joints are often not reliable and, hence they are not used in our method. Thus, there are 15 joints 14 limbs. The joint Head is considered as the origin of the 15 points. For each limb, we compute a unit difference vector between its two joints:
\begin{equation}
(\Delta x_{ij}, \Delta y_{ij}, \Delta z_{ij}) = \dfrac{(x_{i}, y_{i}, z_{i}) - (x_{j}, y_{j}, z_{j})}{d_{ij}}
\end{equation}
where $i$ and $j$ represent the current joint and reference joint, respectively; $d_{ij}$ is the Euclidean distance between the two joints. For example, as illustrated in Fig. 1, to compute the orientation of the limb between joint Hand Right and Wrist Right (highlighted in red), the Wrist Right joint is regarded as the sphere center and Eq. (1) is used to compute the unit difference vector.

Each element of the unit difference vector is quantized into three states: $-1$, $0$ and $1$. If $|\Delta x_{ij}|$ $\leq$ $threshold$ then $q(\Delta x_{ij})= 0$; if $\Delta x_{ij}$ $>$ $threshold$ then $q(\Delta x_{ij}) = 1$; else $q(\Delta x_{ij}) = -1$. Thus, there are 27 possible states for each unit difference vector, and each state is encoded as one element of a feature vector, so the dimension of the feature vector for each pose is 14$\times$27 = 378 after concatenating all feature vectors of the 14 limbs. For each element of the feature vector, if the corresponding orientation between two joints is bid to one state, then the relative position is labelled to 1, otherwise, it is 0. Therefore, the feature vectors are very sparse, only 14 positions in each feature vector are 1 (not zeros). The threshold is an empirical value which is dependent on the noise characteristics of the skeleton data.

For each frame of skeleton, a quantized 378 dimensional feature vector is calculated as described above. This feature vector is reduced to a 14 dimensional feature vector with each element being the index to a non-zero element of the 378-dimensional feature vector.

To extract mid-level features for action representation, the 14 limbs are combined into 7 body parts. As illustrated in Fig. 1, the dotted line contains joints Hand Right, Wrist Right and Elbow Right, and these three limbs form one part. In this way, seven body parts are formed, namely, Head-Shoulder Center, Should Center-Shoulder Left-Elbow Left-Wrist Left, Shoulder Center-Shoulder Right-Elbow Right-Wrist Right, Shoulder Center-Hip Center-Hip Left, Hip Left-Knee Left-Angle Left,Shoulder Center-Hip Center-Hip Right and Hip-Right-Knee Right-Angle Right. According to the Degree of Freedom (DoF) of joints \cite{zatsiorsky1998kinematics}, each body part is encoded with different number of states and the total number of states is denoted as $NDF$, which is currently an empirical parameter. It should be adjusted according to the complexity of the actions to be recognized and noise level of the dataset.   

To explore temporal information and keep motion information at the same time after frequent data mining (generally, frequent data mining can only mine the most frequent pattens which can not be guaranteed as discriminative patterns), a novel way is proposed. Seven states for each frame will be obtained after combination, and the unique states of continuous $C$ frames, as illustrated in Fig. 1, where $C = 3$, are counted and form a new mid-level feature vector, denoted as $\{f_{i}|i = 1, ..., n_{A}\}$. This new feature vector contains both pose information of the current frame and the motion information in the continuous $C$ frames, because the repeated states in the continuous frames can be regarded as static pose information and the different ones with other frames can capture the motion information. This feature vector is used to build transactions described in the next section. The pattens after mining can be the combinations of several body parts in different frames, thus the temporal order information can be easily maintained. 

\subsection{Building Transactions}

Each instance of action $A$ is represented by a set of above mid-level features $\{f_{i}|i = 1, ..., n_{A}\}$ and a class label $c$, $c \in \{1...C\}$. The set of features for all the action samples is denoted by $\Omega$. The dimensionality of the feature vector is denoted as $W$ and in our case $|$W$|$ $\geq 7$.
\subsubsection{Items, Transactions and Frequencies}
Each element in a feature vector for continuous $C$ poses is defined as an item, and an item is denoted as $\omega$, where $\omega$ $\in$ $(0, NDF]$ and $\omega$ $\in$ $\mathbb{N}$.

The set of \textit{transactions} $X$ from the set $\Omega$ is created next. For each $\textbf{x}$ $\in$ $\Omega$ there is one transaction $x$ (i.e. a set of items). This transaction $x$ contains all the items $\omega_{j}$. A \textit{local pattern} is an itemset $t \subseteq \Gamma$, where $\Gamma$ represents the set of all possible items. For a local pattern $t$, the set of transactions that include the pattern $t$ is defined as: $X(t) = \{x \in X|t \subseteq x\}$. The \textit{frequency} of $t$ is $|X(t)|$, also known as the \textit{support} of the pattern $t$ or $supp(t)$.

\subsubsection{Frequent Local Part} 
For a given constant $T$, also known as the minimum support threshold, a local pattern $t$ is \textit{frequent} if $supp(t) \geq T$. A pattern $t$ is said to be \textit{closed} if there exists no pattern $t^{'}$ that $t \subset t^{'}$ and $supp(t) = supp(t^{'})$. The set of frequent closed patterns is a compact representation of the frequent patterns, and such a frequent and closed local part pattern is referred to as \textit{Frequent Local Part} of \textit{FLP}.

\subsection{Mining \& Selecting Relevant FLPs}

\subsubsection{FLPs Mining} Given the set of transaction $X$, any existing frequent mining algorithm can be used to find the set of \textit{FLPs} $\Upsilon$. In our work, the optimised \textit{LCM} algorithm \cite{Uno03lcm2003} is used as in \cite{fernando2014mining}. \textit{LCM} uses a \textit{prefix preserving closure extension} to completely enumerate closed itemsets.

\subsubsection{Encoding a New Action with FLPs} Given a new action, the features can be extracted according to the section $A$ and each feature vector can be converted into a transaction $x$ and for each \textit{FLP} pattern $t$ $\in$ $\Upsilon$ it can be checked whether $t \subseteq x$. If $t \subseteq x$ is true, then $x$ is an \textit{instance} of the \textit{FLP} pattern $t$. The frequency of a pattern $t$ in a given action $A_{j}$ (i.e. the number of instances of $t$ in $A_{j}$) is denoted as $F(t|A_{j})$.

\subsubsection{Selecting the Best FLPs for Action Recognition}
The \textit{FLPs} set $\Upsilon$ is considered as a candidate set of mid-level features to represent an action. Therefore, the most useful \textit{FLP} patterns from $\Upsilon$ is needed to be selected because $i)$ the number of generated \textit{FLP} patterns is huge and $ii)$ not all discovered \textit{FLP} patterns are equally important to the action recognition task. Usually, relevant patterns are those \textit{discriminative} and \textit{non-redundant}. On top of that, a new criterion, \textit{representativity} is also used. As a result, some patterns may be frequent and appear to be discriminative but they may occur in very few actions (e.g. noise pose). Such features are not representative and therefore not the best choice for action recognition. A good \textit{FLP} pattern should be at the same time discriminative, representative and non-redundant. In this section, how to select such patterns is discussed.

The methods used in \cite{fernando2014mining} are followed to find the most suitable pattern subset $\chi$, where $\chi \subset \Upsilon$, for action recognition. To do this the \textit{gain} of a pattern $t$ is denoted by $G(t)$ (s.t. $t \not\in \chi $ and $t \in \Upsilon$) and defined as follows:
\begin{equation}
G(t) = S(t) - \max_{s\in \chi}\{R(s,t)\cdot \min (S(t), S(s))\}
\end{equation}
where $S(t)$ is the overall relevance of a pattern $t$ and $R(s,t)$ is the redundancy between two patterns $s$, $t$. In Eq. (2), a pattern $t$ has a higher gain $G(t)$ if it has a higher relevance $S(t)$ (i.e. it is discriminative and representative) and if the pattern $t$ is non redundant with any pattern $s$ in set $\chi$ (i.e. $R(s,t)$ is small).
 $S(t)$ is defined as:
\begin{equation}
S(t) = D(t) \times O(t),
\end{equation} 
and $R(s,t)$ is defined as:
\begin{equation}
\begin{split}
R(s,t) = &\exp\{-[p(t)\cdot D_{KL}(p(A|t)||p(A|\{t,s\}))\\ &+ p(s)\cdot D_{KL}(p(A|s)||p(A|\{t,s\}))]\}.
\end{split}
\end{equation}
Following a similar approach in \cite{yan2005summarizing} to find affinity between patterns, two patterns $t$ and $s$ $\in$ $\Upsilon$ are redundant if they follow similar document distributions, i.e. if $p(A|t) \approx p(A|s) \approx p(A|\{t,s\})$ where $p(A|\{t,s\})$ is the document distribution given both patterns $\{t,s\}$.

In Eq. (3), $D(t)$ is the \textit{discriminability score}. Following the entropy-based approach in \cite{cheng2007discriminative}, and a high value of $D(t)$ implies that the pattern $t$ occurs only in very few actions; $O(t)$ is the \textit{representativity score} for a pattern $t$  and it considers the divergence between the optimal distribution for class $c$ $p(A|t_{c}^{\ast})$ and the distribution for pattern $t$ $p(A|t)$, and then takes the best match over all classes. The optimal distribution is such that $i)$ the pattern occurs only in actions of class $c$, i.e. $p(c|t_{c}^{\ast}) = 1$ (giving also a discriminability score of 1), and $ii)$ the pattern instances are equally distributed among all the actions of class $c$, i.e. $\forall A_{j}, A_{k}$ in class $c$, $p(A_{j}|t_{c}^{*}) = p(A_{k}|t_{c}^{*}) = (1/N_{c})$ where $N_{c}$ is the number of samples of class $c$. An optimal pattern, denoted by $t_{c}^{*}$ for class $c$, is a pattern which has above two properties. 

The \textit{discriminability score} and \textit{representativity score} are defined as:
\begin{equation}
D(t) = 1 + \dfrac{\sum_{c} p(c|t)\cdot \log p(c|t)}{\log C},
\end{equation}
\begin{equation}
O(t) = \max\limits_{c}(\exp\{-[D_{KL}(p(A|t_{c}^{\ast})||p(A|t))]\})
\end{equation}
where $p(c|t)$ is the probability of class $c$ given the pattern $t$, computed as follows:
\begin{equation}
p(c|t) = \frac{\sum_{j=1}^{N} F(t|A_{j})\cdot p(c|A_{j})}{\sum_{j=1}^{N} F(t|A_{j})};
\end{equation}
$D_{KL}(.||.)$ is the Kullback-Leibler divergence between two distributions; $p(A|t)$ is computed empirically from the frequencies $F(t|A_{j})$ of the pattern $t$:
\begin{equation}
p(A|t) = \dfrac{F(t|A)}{\sum_{j} F(t|A_{j})}
\end{equation}
Here, $A_{j}$ is the $j^{th}$ action and $N$ is the total number of actions in the dataset. $p(c|A) = 1$ if the class label of $A_{j}$ is $c$ and $0$ otherwise; $p(c|t_{c}^{\ast})$ is the optimal distribution with respect to a class $c$.\\
In Eq. (4), $p(t)$ is the probability of pattern $t$ and it is defined as:
\begin{equation}
p(t) = \dfrac{\sum_{A_{j}} F(t|A_{j})}{\sum_{t_{j} \in \Upsilon}\sum_{A_{j}}F(t_{j}|A_{j})}
\end{equation}
while $p(A|\{t,s\})$ is the document distribution given both patterns $\{t,s\}$ and it is defined as:
\begin{equation}
p(A|\{t,s\}) = \dfrac{F(t|A) + F(s|A)}{\sum_{j} F(t|A_{j}) + F(s|A_{j})}
\end{equation}
To find the best $K$ patterns the following greedy process is used. First the most relevant pattern is added to the relevant pattern set $\chi$. Then the pattern with the highest gain (non redundant but relevant) is searched out and this pattern is added into the set $\chi$ until $K$ patterns are added (or until no more relevant patterns can be found). For more detailed discussions, \cite{fernando2014mining} is recommended to refer to.

\subsection{Building Bag-of-FLPs \& Classification}
After computing the $K$ most relevant and non-redundant \textit{FLPs}, each action can be represented by a new representation called \textit{bag-of-FLPs} by counting the occurrences of such \textit{FLPs} in the action. Let $L$ be such a \textit{bag-of-FLPs} for action $A_{L}$ and $M$ be the \textit{bag-of-FLPs} for action $A_{M}$.

An SVM \cite{CC01a} is trained to classify the actions. The SVM uses the following kernel to calculate the similarities between the \textit{bag-of-FLPs} of $L$ and $M$.
\begin{equation}
K(L,M) = \sum\limits_{i}\min (\sqrt{L(i)},\sqrt{M(i)})
\end{equation}
 Here $L(i)$ is the frequency of the $i^{th}$ selected pattern in histogram $L$. It is a standard histogram intersection kernel with non-linear weighting. This reduces the importance of highly frequent patterns and is necessary since there is a large variability in pattern frequencies. 
\section{Experimental Results}
Two benchmark datasets, MSR-DailyActivity3D \cite{Wang2012} and MSR-ActionPairs3D \cite{Oreifej2013}, were used to evaluate the proposed method and the results are compared with those reported in other papers on the same datasets and under the same training and testing configuration.
\subsection{Experimental Setup} 
In our method, there are several parameters that need to be tuned, the \textit{threshold} $T$, the number of states $NDF$, the number of relevant patterns $K$, the continuous frames $C$, minimum support $S$ and maximum support $U$. For different datasets, different sets of parameters were learned through cross-validation to optimise the performance. Specifically, two-third of the entire training dataset was used as training and the rest one-third was used for validation to tune the parameters. The ranges of the parameters are empirical. In general, the threshold $T$ is dependent on the noise level of the dataset. The higher the noise the larger its value. This is an important parameter because it affects the states of limbs computed from the skeleton data. However, such sensitivity can be reduced by setting a large number, $NDF$ (i.e. over 600) of states. The number of relevant patterns $K$ is dependent on the complexity of the actions to be recognized, the more actions in the dataset, the larger number it should be. The number of continuous frames $C$ is affected by the complexity of required temporal information to encode the actions. If the dataset has pair actions, for example, two actions of each pair are similar in motion (have similar trajectories) and shape (have similar objects), the value of $C$ should be large. However, a large $C$ leads to high memory and post-processing requirement. The values of the minimum support $S$ and maximum support $U$ effect the number of generated patterns before pattern selection. We observed that if $S$ is large, $U$ should also be large; If $S$ is small, $U$ should also be small. Generally, $S$ and $U$ are set to reduce the computational time for post-processing. In fact, there are many combinations of these two parameters to get the best results. In the other words, the performance of the proposed method is not much sensitive to the choice of $S$ and $U$.
\subsection{MSR DailyActivity3D} 
The MSR DailyActivity3D dataset consists of 10 subjects and 16 activities: \textit{drink, eat, read book, call cellphone, write on a paper, use laptop, use vacuum cleaner, cheer up, sit still, toss paper, play game, lay down on sofa, walk, play guitar, stand up, sit down}. Fig. 3 shows some sample frames for the activities. 
\begin{figure}[!ht]
\begin{center}
{\includegraphics[height = 70mm, width = 90mm]{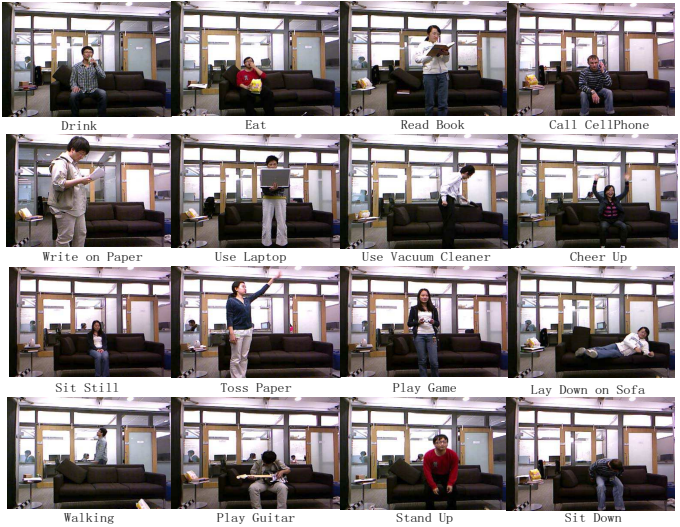}}
\end{center}
\caption{Sample frames of the MSR DailyActivity3D dataset.}
\label{fig:framework}
\end{figure}
Each subject performed each activity twice, once in standing position and once in sitting position. In total, there are 320 samples. This dataset has large intra-class variations and involves human-object interactions, which is challenging for recognition only by 3D joints. Experiments were performed based on cross-subject test setting described in \cite{Zanfir_movpose}, i.e. five subjects (1, 2, 3, 4, 5) were used for training and the rest 5 subjects were used for testing. Table I shows the results of our methods compared with other published results. 
\begin{table}[!hbp]
\centering
\caption{Comparison on MSR-DailyActivity Dataset}
\begin{tabular}{|c|c|}
\hline
Methods & Accuracy (\%)\\
\hline
Dynamic Temporal Warping \cite{muller2006motion} & 54.0\\
\hline
Moving Pose \cite{Zanfir_movpose} & 73.8  \\
\hline
Actionlet Ensemble on Joint Features \cite{wang2014learning} & 74.0  \\
\hline
Proposed Method & 78.8  \\
\hline
\end{tabular}
\end{table}
For this dataset, $T = 0.15, NDF = 600, K = 30000, C = 3, S = 15, U = 180$.  
As seen, although this dataset is quite challenging, our method obtained promising results based only on skeleton data. The confusion matrix is illustrated in Fig. 4. 
\begin{figure}[!ht]
\begin{center}
{\includegraphics[height = 50mm, width = 90mm]{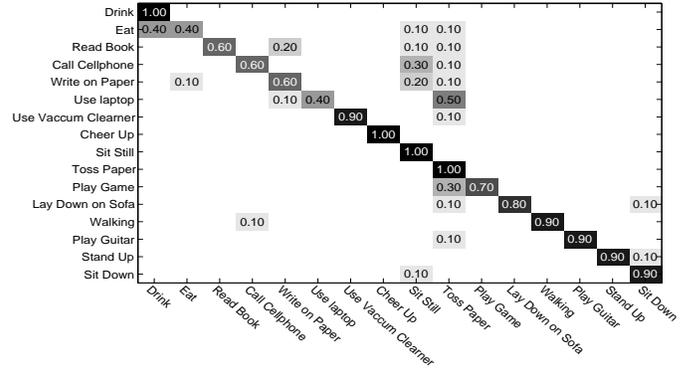}}
\end{center}
\caption{The confusion matrix of our proposed method for MSR-DailyActivity3D.}
\label{fig:framework}
\end{figure}
From the confusion matrix, it can be seen that activities such as ``Drink", ``Cheer Up", ``Sit Still", ``Toss Paper" are relatively easy to recognise, while ``Eat" and ``Use laptop" are relatively difficult to recognise. The reason for the difficulties is that for these human-object interactions, object information was not available from skeleton data which makes these interactions are almost the same in terms of motion reflected in the skeleton data.

\subsection{MSR ActionPairs3D}
The MSR ActionPairs3D dataset \cite{Oreifej2013} is a paired-activity dataset captured by a Kinect camera. This dataset contains 12 activities (i.e. six pairs) of 10 subjects with each subject performing each activity 3 times. The pair actions are: Pick up a box/Put down a box, Lift a box/Place a box, Push a chair/Pull a chair, Wear a hat/Take off hat, Put on a backpack/Take off a backpack, Stick a poster/Remove a poster. Some sample frames for the activities of this dataset are shown in Fig. 5.
\begin{figure}[!ht]
\begin{center}
{\includegraphics[height = 50mm, width = 90mm]{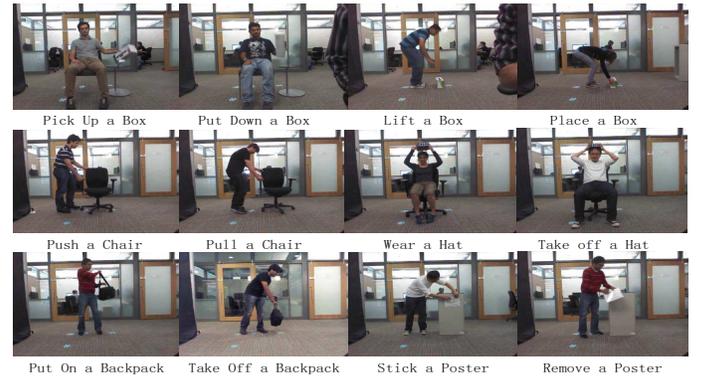}}
\end{center}
\caption{Sample frames of the MSR ActionPairs3D dataset.}
\label{fig:framework}
\end{figure} 
This dataset is collected to investigate how the temporal order affects activity recognition.
\begin{table}[!hbp]
\centering
\caption{Comparison on MSR-ActionPairs Dataset}
\begin{tabular}{|c|c|}
\hline
Methods & Accuracy (\%)\\
\hline
Skeleton + LOP \cite{Wang2012} & 63.33  \\
\hline
Depth Motion Maps \cite{yang2012recognizing} & 66.11\\
\hline
Proposed Method & 75.56  \\
\hline
\end{tabular}
\end{table}
Experiments were set to the same configuration as \cite{Oreifej2013}, namely, the first five actors are used for testing, and the rest for training. For this dataset, $T = 0.11, NDF = 1000, K = 10000, C = 4, S = 3, U = 100$. We compare our performance in this dataset with two methods whose results were reported in \cite{Oreifej2013}. Table II shows the comparisons with other methods tested on this dataset.  

The confusion matrix is shown in Fig. 6. 
\begin{figure}[!ht]
\begin{center}
{\includegraphics[height = 50mm, width = 90mm]{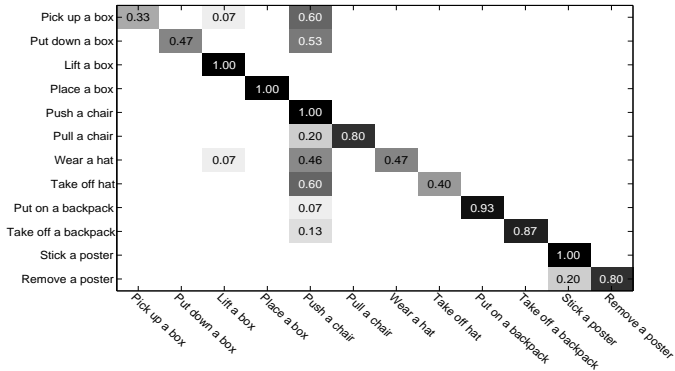}}
\end{center}
\caption{The confusion matrix of our proposed method for MSR-ActionPairs3D.}
\label{fig:framework}
\end{figure}
From the confusion matrix, it can be seen that activities such as ``Lift a box", ``Place a Box", ``Push a Chair", ``Stick a Poster" are easy for our method to recognise, while ``Pich up a Box" and ``Take off Hat" are relatively difficult to recognise. The results have verified that our method can distinguish temporal orders in actions, however, it still can be confused with other actions which were not paired. One possible reason for causing the confusion between some actions, for instance, ``Pick up a Box" and ``Push a Chair", is the 3-state quantization of the unit different vectors. This issue can be addressed by quantizing the vector into more states.
\section{Conclusion} 
In this paper, a new representation is proposed and effective data mining method is adopted to mine the mid-level patterns (different compositions of body parts) for action recognition. A novel method to explore temporal information and mine the different combinations of different body parts in different frames is proposed. The strength of the proposed method has been demonstrated through the state-of-the-art results obtained on the recent and challenging benchmark datasets for activity and action recognition. However, the proposed method can be further improved by combining depth or RGB data to explore the human-object interactions. With the increasing popularity of Kinect-based action recognition and data mining methods in computer vision, the proposed method has promising potentialities in practical applications.


\bibliographystyle{IEEEtran}
\bibliography{References}

\end{document}